\documentclass[10pt,twocolumn,letterpaper]{article}

\usepackage{cvpr}
\usepackage{times}
\usepackage{epsfig}
\usepackage{graphicx}
\usepackage{amsmath}
\usepackage{amssymb}
\usepackage{booktabs, algorithm2e}
\usepackage{mdframed}

\mdfsetup{
    linewidth=0.6pt
}

\usepackage[xcdraw]{xcolor}

\newcommand{\ts}{\textsuperscript}

\usepackage[pagebackref=true,breaklinks=true,letterpaper=true,colorlinks,bookmarks=false]{hyperref}

\cvprfinalcopy % *** Uncomment this line for the final submission

\begin{document}

\title{Crypto-Oriented Neural Architecture Design}

\author{Avital Shafran \qquad Gil Segev \qquad Shmuel Peleg \qquad Yedid Hoshen\\
School of Computer Science and Engineering\\
The Hebrew University of Jerusalem, Israel\\}

\maketitle

\begin{abstract}
As neural networks revolutionize many applications, significant privacy conflicts between model users and providers emerge. The cryptography community developed a variety of techniques for secure computation to address such privacy issues. As generic techniques for secure computation are typically prohibitively ineffective, many efforts focus on optimizing their underlying cryptographic tools. Differently, we propose to optimize the initial design of crypto-oriented neural architectures and provide a novel Partial Activation layer. The proposed layer is much faster for secure computation. Evaluating our method on three state-of-the-art architectures (SqueezeNet, ShuffleNetV2, and MobileNetV2) demonstrates significant improvement to the efficiency of secure inference on common evaluation metrics.

\end{abstract}

\section{Introduction}
\label{sec:introduction}

Deep neural networks are revolutionizing many applications, but practical use may be slowed down by privacy concerns. As an illustrative example, let us consider a hospital that wishes to use an external diagnosis service for its medical images (e.g.\ MRI scans). In some cases, the hospital would be prevented from sharing the medical data of its patients for privacy reasons. On the other hand, the diagnosis company may not be willing to share its model with the hospital to safeguard its intellectual property. Such privacy conflicts could prevent hospitals from using neural network services for improving healthcare. The ability to evaluate neural network models on private data will allow the use of neural network services in privacy-sensitive applications.

The privacy challenge has attracted significant research in the cryptography community. Cryptographic tools were developed to convert any computation to secure computation, i.e.\ computation where the view of each involved party is guaranteed not to reveal any non-essential information on the inputs of the other parties. The deep learning setting consists of two parties, one providing the data and the other providing the neural network model. Secure computation is typically slower then non-secure computation and requires much higher networking bandwidth. 
Recently, various approaches were proposed for secure computation of neural networks \cite{bae2018security, tanuwidjaja2019survey}. However, due to the efficiency limitations of secure computation, these approaches have so far been somewhat limited to simple architectures, decreasing their accuracy and applicability.

\paragraph{Our Contribution.} Instead of using existing architectures and optimizing the cryptographic protocols, we take a complementary approach. We propose to design new neural network architectures that are crypto-oriented.  We propose a novel \textit{partial activation} layer for the design of crypto-oriented neural architectures. Non-linear activations such as ReLU are very expensive for secure computation. We propose to split each layer to two branches, applying the non-linear activation on one branch only, to significantly reduce required resources. Different activation layers have different effects on accuracy, therefore we propose to carefully choose the ratio between the branches in different layers. For layers whose removal makes no significant impact on accuracy we suggest to use a $0\%$-partial activation layer, i.e.\ removing the activation layer completely. For clarity, we will refer to this as "activation layer removal" rather than $0\%$-partial activation.

Given our proposed \textit{partial activation} layer, we present new crypto-oriented architectures based on three popular (non crytpo-oriented) efficient neural network architectures: MobileNetV2 \cite{sandler2018mobilenetv2}, ShuffleNetV2 \cite{ma2018shufflenet} and SqueezeNet \cite{iandola2016squeezenet}. Our new architectures are significantly more efficient than their non-crypto-oriented counterparts, with a minor loss of accuracy.

\begin{table*}
\centering
    \begin{tabular}{lcccc|cccc}
    \toprule
    &
    \multicolumn{4}{c|}{\textbf{CIFAR-10 /\ 100}} &
    \multicolumn{4}{c}{\textbf{MNIST /\ FASHION}}  \\
    \midrule
    \midrule
    \textbf{Model} & 
    \textbf{Accuracy} & \textbf{Comm.}  & \textbf{Rounds} & \textbf{Runtime}
    & \textbf{Accuracy} & \textbf{Comm.}  & \textbf{Rounds} & \textbf{Runtime}\\
    & \scriptsize{CIFAR10 /\ 100} & \textbf{(MB)} & & \textbf{(sec)}
    & \scriptsize{MNIST /\ FASHION} & \textbf{(MB)} & & \textbf{(sec)}\\
    \midrule
    Squeeze-orig 
    & 92.49 /\ 70.41 & 327.2 & 393 & 14.59 
    & 99.27 /\ 94.05 & 248.07 & 393 & 14.36\\
    Squeeze-ours
    & 91.87 /\ 69.7 & 149.59 & 232 & 9.03
    & 99.08 /\ 93.29 & 66.77 & 152 & 6.34\\
    \midrule
    Shuffle-orig 
    & 92.6 /\ 70.95 & 311.63 & 484 & 24.37
    & 99.26 /\ 93.51 & 249.36 & 484 & 23.8 \\
    Shuffle-ours 
    & 92.5 /\ 70.07 & 157.63 & 294 & 14.88 
    & 99.23 /\ 93.4 & 104.3 & 294 & 11.4\\
    \midrule
    Mobile-orig 
    & 94.49 /\ 74.8 & 1926.34 & 806 & 41.01
    & 99.23 /\ 94.51& 1517.37 & 806 & 38.41 \\
    Mobile-ours 
    & 93.44 /\ 72.61 & 403.52 & 296 & 17.11 
    & 99.25 /\ 94.29& 250.42 & 296 & 16.12 \\
	\bottomrule
    \end{tabular}

\caption{Comparison of performance on secure classification using a few known networks (SqueezeNet, ShuffleNetV2, and MobileNetV2), before and after our proposed crypto oriented modifications. Our modification provides substantial increase of efficiency with a minor reduction of accuracy. While the accuracy is noted separately for each dataset, the complexity measures for the two CIFAR datasets  are almost the same, as well as for the two MNIST datasets.}
\label{tab:crypto_oriented_results}
\vspace{-0.3cm}
\end{table*}

%-------------------------------------------------------------------------
%-------------------------------------------------------------------------
\section{Background}
\label{sec:background}

\subsection{Privacy-Preserving Machine Learning}

Research on privacy-preserving machine learning has so far focused on two main challenges: Privacy-preserving training and privacy-preserving inference. Privacy-preserving training \cite{shokri2015privacy, abadi2016deep, bonawitz2017practical} aims at enabling neural networks to be trained with private data. This happens, for example, when private training data arrives from different sources, and data privacy must be protected from all other parties. 

In this work we address the challenge of privacy-preserving inference. A pre-trained neural network is provided, and the goal is to transform the network to process (possibly interactively) encrypted data. The network's output should also be encrypted, and only the data owner can decode it. This enables users with private data, such as medical records, to rely on the services of a model provider. 

Existing privacy-preserving inference methods \cite{bae2018security, tanuwidjaja2019survey} rely on three cryptographic approaches, developed by the cryptography community in the context of secure computation: Homomorphic encryption, garbled circuits, and secret sharing.

Given a neural network $N$ with depth $k$, it can be represented by a list of composed layers:
\begin{equation}
N(X)= F_k \circ F_{k-1} \circ ... \circ F_1(X)
\vspace{-0.09cm}
\end{equation}
where $F_i$ is the $i\ts{th}$ layer of the network, and $X$ is the input to the network. Using the above cryptographic tools, each layer can be transformed into a privacy-preserving layer $\hat{F_i}$ such that given the encoding  $\hat{X}$ of a private input $X$ the output of:
% \vspace{-0.2cm}
\begin{equation}
\widehat{N(X)} = \hat{F}_k \circ \hat{F}_{k-1} \circ ... \circ \hat{F}_1(\hat{X})
\vspace{-0.09cm}
\end{equation} 
is encrypted as well, and can be decoded only by the owner of $X$ to compute $N(X)$. 

\paragraph{The challenge of practicality.} Despite the extensive research within the cryptography community towards more practical secure computation protocols, the above approaches are practical mainly for simple computations. In particular, homomorphic encryption and secret sharing are most suitable for layers that correspond to affine functions (or to polynomials of small constant degrees). Non-affine layers (e.g. ReLU or Max Pooling) lead to significant overhead, both in computation and in communication. Garbled circuits can be efficient for layers corresponding to functions that can be represented via small Boolean circuits, but interaction between the parties for computing every layer is required. This may be undesirable in many scenarios. 

\paragraph{Homomorphic encryption.} Homomorphic encryption \cite{gentry2009fully, brakerski2014leveled} allows to compute an arbitrary function $f$ on an encrypted input, without decryption or knowledge of the private key. In other words, for every function $f$ and encrypted input  $\hat{x} = enc(x)$ it is possible to compute an encryption $\widehat{f(x)}$ of $f(x)$ without knowing the secret key that was used to encrypt $x$. Gilad-Bachrach et al.\ \cite{gilad2016cryptonets} relied on homomorphic encryption in their CryptoNets system, replacing the ReLU activation layers with square activation. However this approach significantly increased the overall inference time. \cite{hesamifard2017cryptodl, chabanne2017privacy, sanyal2018tapas, chou2018faster, bourse2018fast} also propose optimization methods using homomorphic encryption.

\paragraph{Garbled circuits.} In the context of layer-by-layer transformations, garbled circuits \cite{yao1986generate} can be roughly viewed as a one-time variant of homomorphic encryption \cite{rouhani2018deepsecure, juvekar2018gazelle, riazi2019xonn}. For two parties, A and B, where A holds a function $f$ (corresponding to a single layer of the network) and B holds an input $x$, the function $f$ is transformed by $A$ into a garbled circuit that computes $f$ on a single encoded input. B will encode its input $x$, and then one of the parties will be able to compute an encoding of $f(x)$ from which $f(x)$ can be retrieved.

\paragraph{Secret sharing.} Secret sharing schemes \cite{Shamir79,Beimel11} provide the ability to share a secret between two or more parties. The secret can be reconstructed by combining the shares of any ``authorized'' subset of the parties (e.g., all parties or any subset of at least a certain size). The shares of any ``unauthorized'' subset do not reveal any information about the secret. As discussed above, secret sharing schemes enable privacy-preserving evaluation of neural networks in a layer-by-layer fashion, where the parties use their shares for all values on each layer  $i$ for computing shares for all value on layer $(i+1)$ \cite{mohassel2017secureml, liu2017oblivious, riazi2018chameleon, wagh2019securenn}.

\subsection{Efficient Neural Network Architecture Design}
Real world tasks require both accuracy and efficiency, sometimes under different constraints e.g.\ hardware. This leads to much work focused on designing deep neural network architectures optimally trading off accuracy and efficiency. 
SqueezeNet \cite{iandola2016squeezenet}, an early approach, reduced the number of model parameters by replacing the commonly used $3 \times 3$ convolutions filters with $1 \times 1$ filters and using squeeze and expand modules. Recent works shifted the focus from reducing parameters to minimizing the number of operations. MobileNetV1 \cite{howard2017mobilenets} utilizes depthwise separable convolution to reduce model complexity and improve efficiency. MobileNetV2 \cite{sandler2018mobilenetv2} further improved this approach by introducing the inverted residual with linear bottleneck block. ShuffleNetV1 \cite{zhang2018shufflenet} relies on pointwise group convolutions to reduce complexity and proposed the \textit{channel shuffle} operation to help information flow across feature channels. ShuffleNetV2 \cite{ma2018shufflenet} proposed guidelines for the design of efficient deep neural network architectures and suggested an improvement over the ShuffleNetV1 architecture. 

\subsection{Efficiency Metrics} Standard neural networks measure efficiency using FLOPs (Floating-Point Operations). Privacy-preserving neural networks require different metrics due to the interactivity introduced by cryptographic protocols. The main measures of efficiency for such protocols are typically their overall communication volume (communication complexity), or the number of rounds of interaction between the parties (round complexity) ~\cite{yao1982protocols, yao1986generate, goldreich1987play, beaver1990round, ishai2000randomizing, franklin1992communication, kushilevitz1997communication, kushelvitz1992privacy, goldwasser1997multi}.

%-------------------------------------------------------------------------

\section{Designing Crypto-Oriented Networks}
\label{sec:method}

\begin{figure}[tb]
\centering

\includegraphics[width=0.99\linewidth, height=0.5\linewidth]{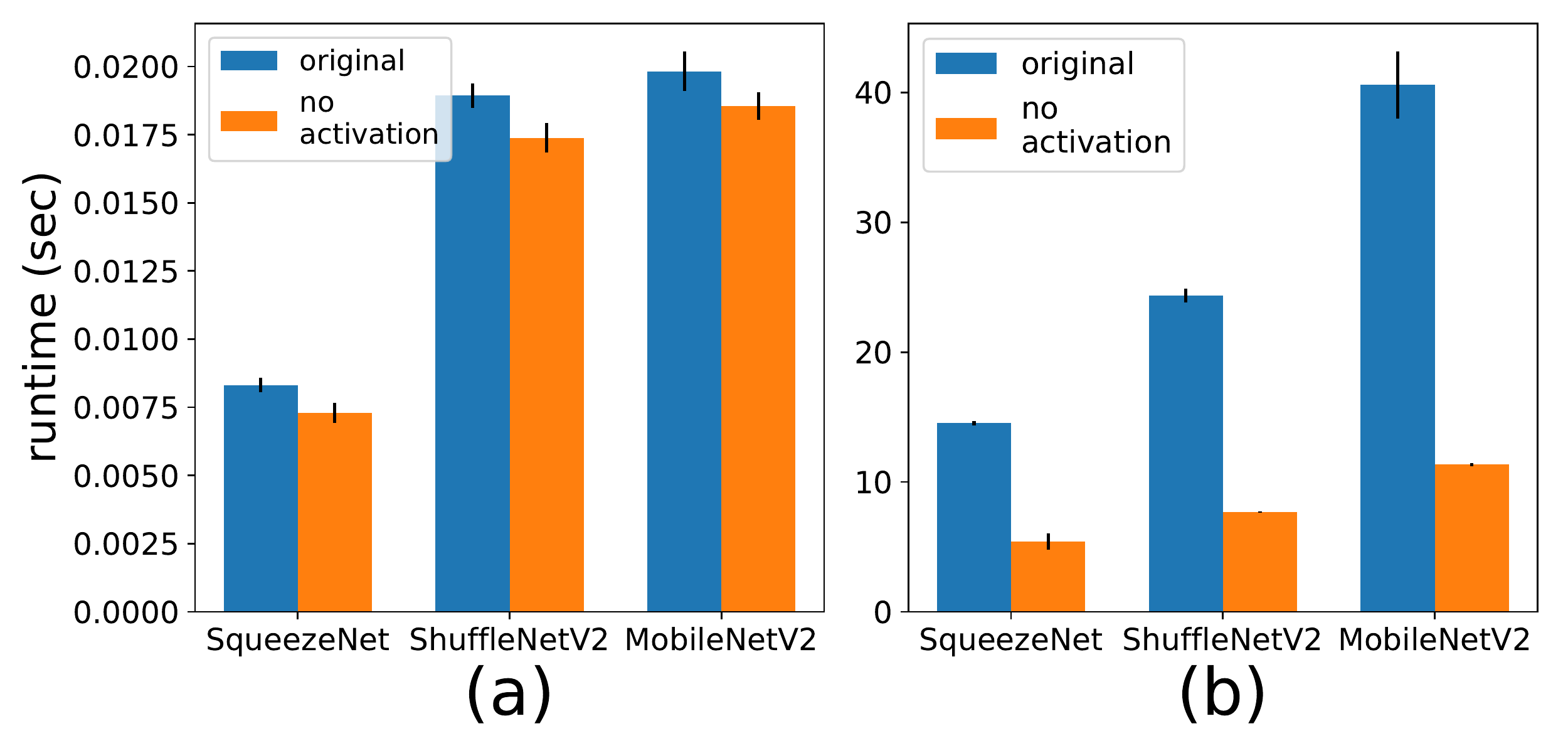}
\caption{Removal of all activation layers has a negligible effect on non-secure inference, but has a drastic reduction of complexity in secure inference. Comparison done on SqueezeNet, ShuffleNetV2 and MobileNetV2 using all datasets. (a) Effects of removal of activation layers on non-secure inference. (b) Effect of removal of activation layers on secure inference. Blue - Original network. Orange - Activation layers removed.}
\label{fig:noReluRuntime}
\vspace{-0.3cm}
\end{figure}

Our goal is to design neural networks that can be computed efficiently in a secure manner for providing privacy-preserving inference mechanisms. We propose a novel \textit{partial activation} layer that exploits the trade-offs that come with the complexity of the cryptographic techniques enabling privacy-preserving inference.

In non-secure computation, the cost of affine operations like addition or multiplication is almost the same as the cost of non-linearities such as maximum or ReLU. As typical neural networks consist of many more additions and multiplications than non-linearities, the cost of non-linearities is negligible~\cite{cong2014minimizing, hunsberger2016training}. Efficient network designs therefore try to limit the number and size of network layers, not taking into account the number of non-linearities.

As explained in Sec.~\ref{sec:background}, the situation is different for privacy-preserving neural networks, as secure computation of non-linearities is much more expensive. Homomorphic encryption methods approximate the ReLU activation with polynomials, and higher polynomial degrees are needed for better accuracy. This comes at the cost of a larger computational complexity. While garbled circuits and secret sharing methods present lighter-weight protocols, they have high communication and round complexities. As a result, the number of non-linearities is an important consideration in the design of efficient privacy-preserving networks. The relative speed of different architectures might change between the secure and non-secure cases, and therefore the optimal architecture in each case may be (and typically is) different.

Fig.~\ref{fig:noReluRuntime} illustrates the remarkable difference between the two scenarios, i.e.\ secure and non-secure inference. We evaluate the inference runtime of three popular architectures - SqueezeNet, ShuffleNetV2 and MobileNetV2 - on the CIFAR-10, CIFAR-100, MNIST and Fashion-MNIST datasets.
We can see that in the secure case, the removal of all activations results in more than a $60\%$ runtime reduction, while in the non-secure case the reduction is negligible - around $10\%$.
This highlights that the number of non-linearities must be taken into account in crypto-oriented neural architecture design.

To obtain an analytic understanding of the relative cost of non-linearity vs. convolution evaluation in privacy-preserving networks, let us consider the analytic cost for a particular protocol, SecureNN \cite{wagh2019securenn}. For a convolution layer, the round and communication complexities for $l$ bit input of size $m \times m \times i$, kernel size $f\times f$ and $o$ output channels is given by 
\begin{gather}
    Rounds_{conv} = 2
    \label{eq:conv_rounds}\\
    Comm_{conv} = (2m^2f^2i + 2f^2oi + m^2o)l
    \label{eq:conv_comm}
\end{gather}
In comparison, the ReLU protocol has a round  and communication complexities of:
\begin{gather}
Rounds_{ReLU} = 10
\label{eq:relu_rounds}\\
Comm_{ReLU} = 8l\log p + 24l
\label{eq:relu_comm}
\end{gather}
where $p$ denotes the field size - each $l$-bit number is secret shared as a vector of $l$ shares, each being a value between $0$ and $p-1$ (inclusive).  

Consider the toy example of a small neural network with input of size $32 \times 32 \times 3$, with a convolution layer with kernel size $3\times 3$ and $16$ output channels followed by a ReLU activation layer. When considering $64$-bit numbers and $p$ equal $67$ (following SecureNN) the convolution layer will require $2$ rounds and $0.58\sf MB$ communication, while the ReLU layer will require $10$ rounds and $9.5\sf MB$ communication -- $5$x more rounds and $16$x more communication. 

In the above, ReLU is only used as an illustration. This applies identically to all other non-linear activation layers such as Leaky-ReLU, ELU, SELU, ReLU6, although the exact numerical trade-offs may differ slightly.

\paragraph{Partial activation layers.}

\begin{figure}[tb]
\centering
\includegraphics[width=0.99\linewidth, height=0.5\linewidth]{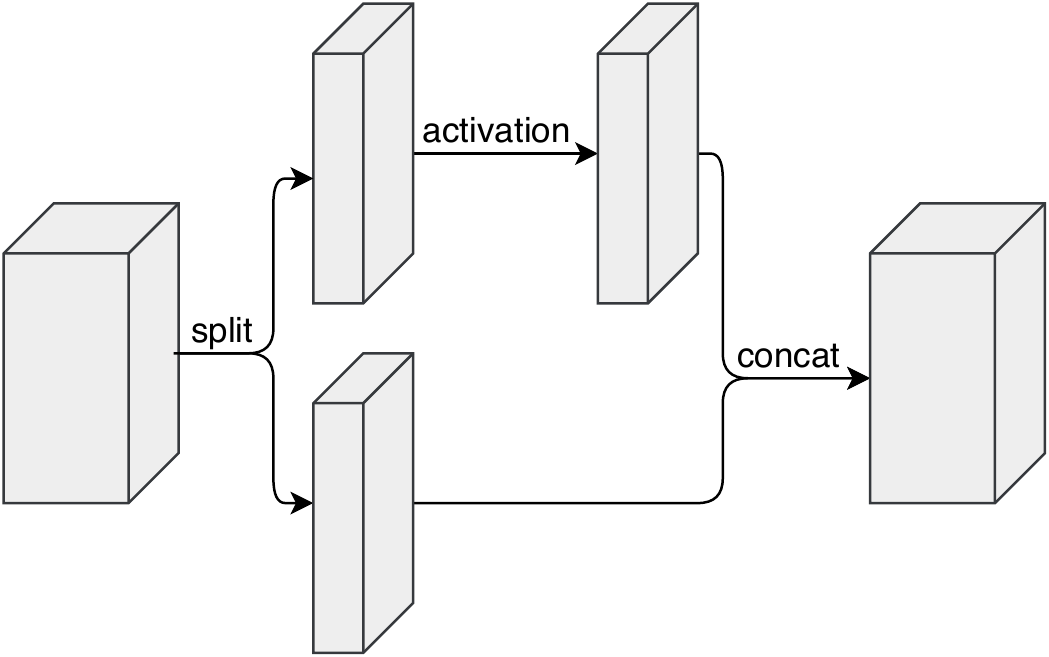}
\caption{In Partial Activation layers, the channels are split and activation is applied only to a subset of the channels, and not applied to the other channels.}
\label{fig:partialActivation}
\vspace{-0.3cm}
\end{figure}

In order to reduce the number of non-linear operations used, we propose a novel \textit{partial activation} layer, illustrated in Fig.~\ref{fig:partialActivation}. \textit{Partial activation} splits the channels into two branches with different ratios, similarly to the \textit{channel split} operation suggested in ShuffleNetV2 \cite{ma2018shufflenet}. The non-linear activation is only applied on one branch. The two branches are then concatenated. By using \textit{partial activation} we can reduce the number of non-linear operations, while keeping the non-linearity of the model. Our experiments show that this operation results in attractive accuracy-efficiency trade-off, dependent on the amount of non-linear channels.  

Beyond reducing the number of non-linearities per layer, it is beneficial to simply remove activations, i.e.\ $0\%$-partial activation, in locations where they do not improve the network accuracy. Dong et al. \cite{dong2017eraserelu} and Zhao et al. \cite{zhao2017training} have studied the effect of erasing some ReLU layers and have shown that this sometimes even improves accuracy. Sandler et al. \cite{sandler2018mobilenetv2} also explored the importance of linear layers and incorporates this notion into the bottleneck residual block.

%

%-------------------------------------------------------------------------

\section{Experiments}

\begin{figure*}[tb]
\centering

\includegraphics[width=0.6\linewidth,height=0.25\linewidth]{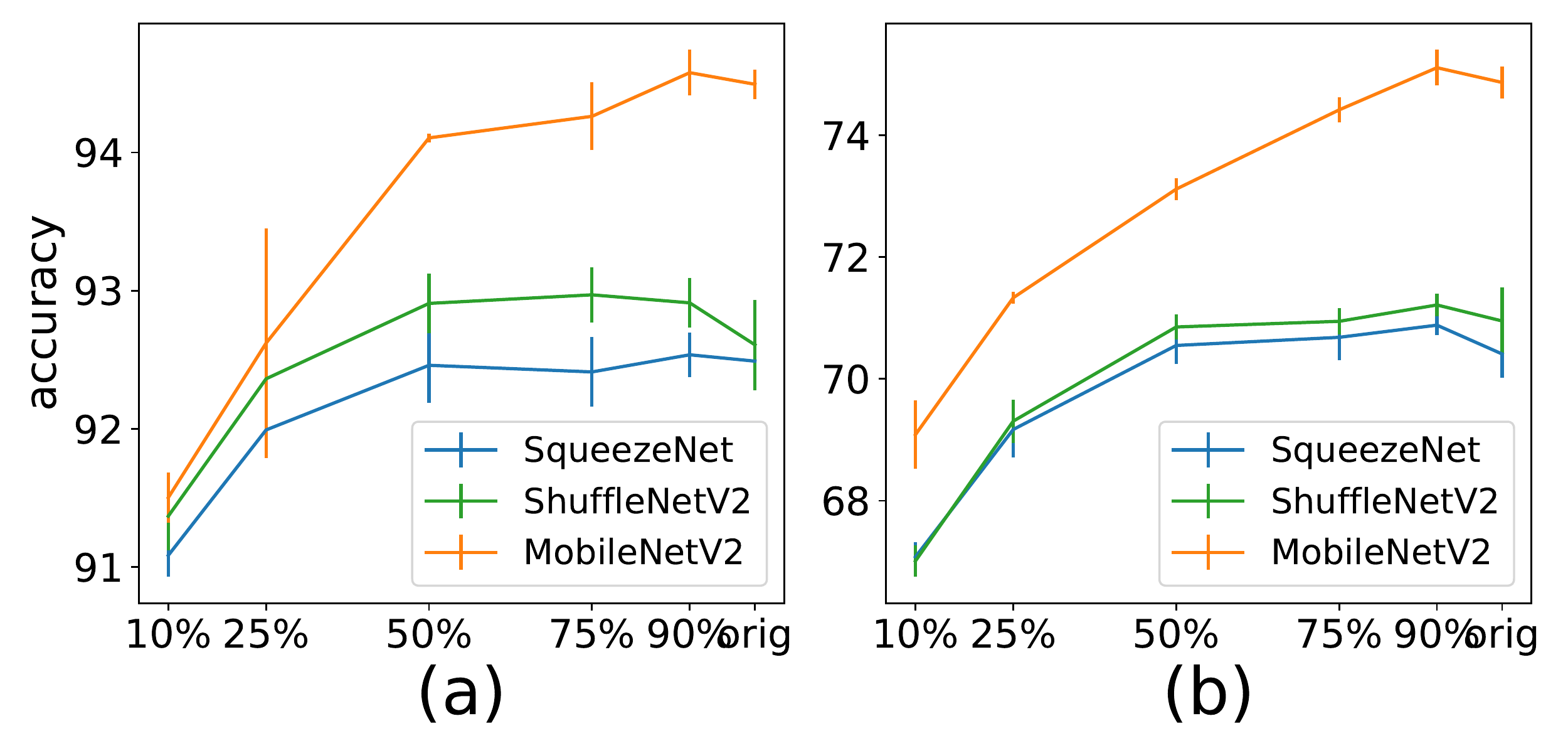}
\includegraphics[width=0.3\linewidth,height=0.25\linewidth]{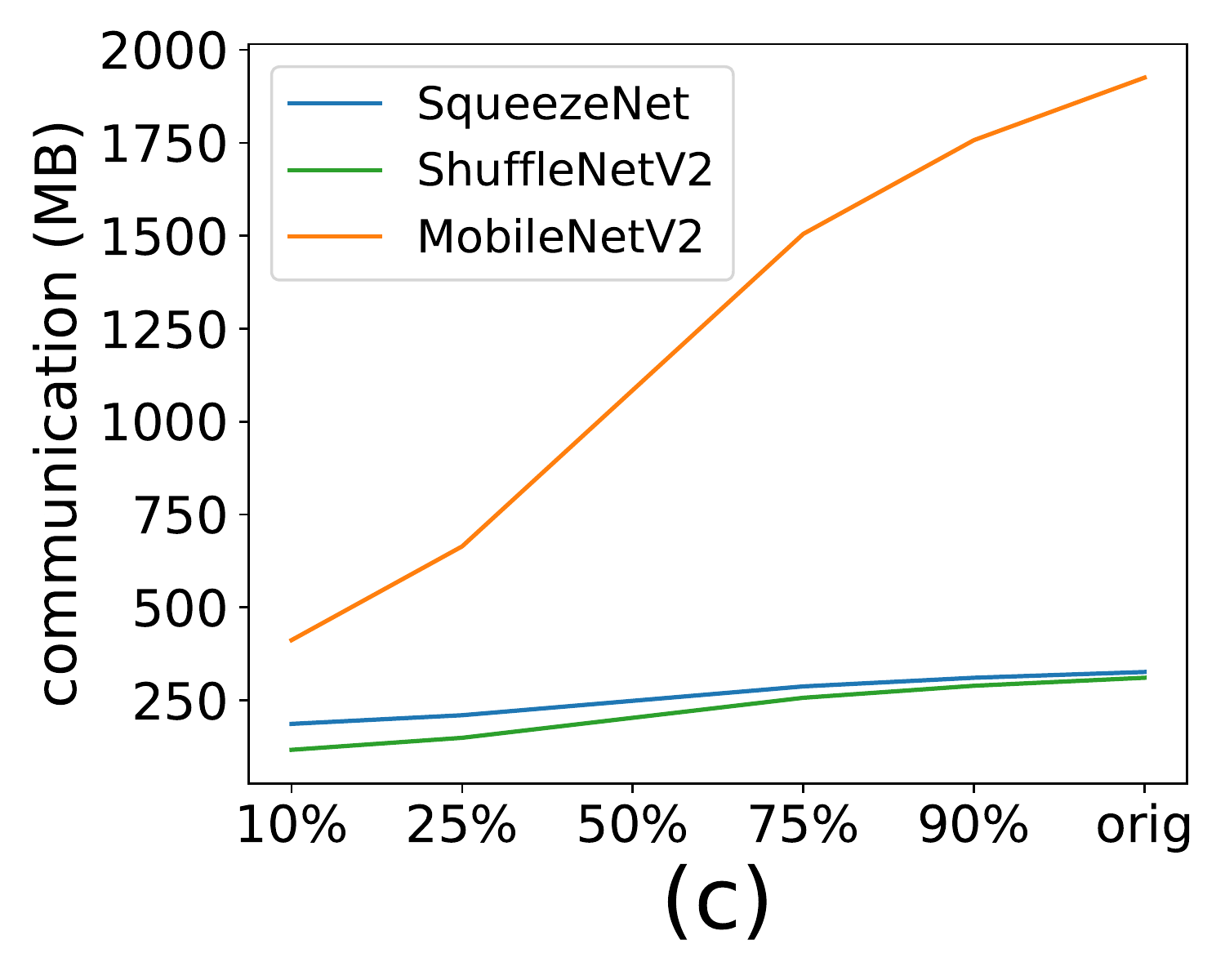}\\
\includegraphics[width=0.8\linewidth,height=0.21\linewidth]{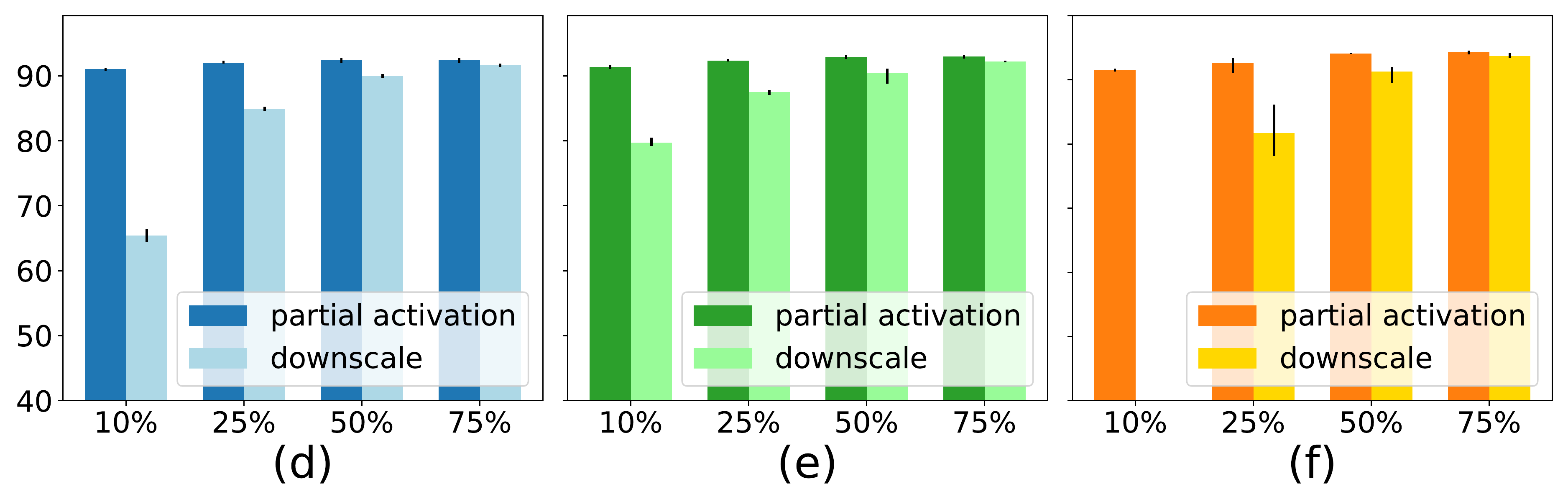}\\
\includegraphics[width=0.8\linewidth,height=0.21\linewidth]{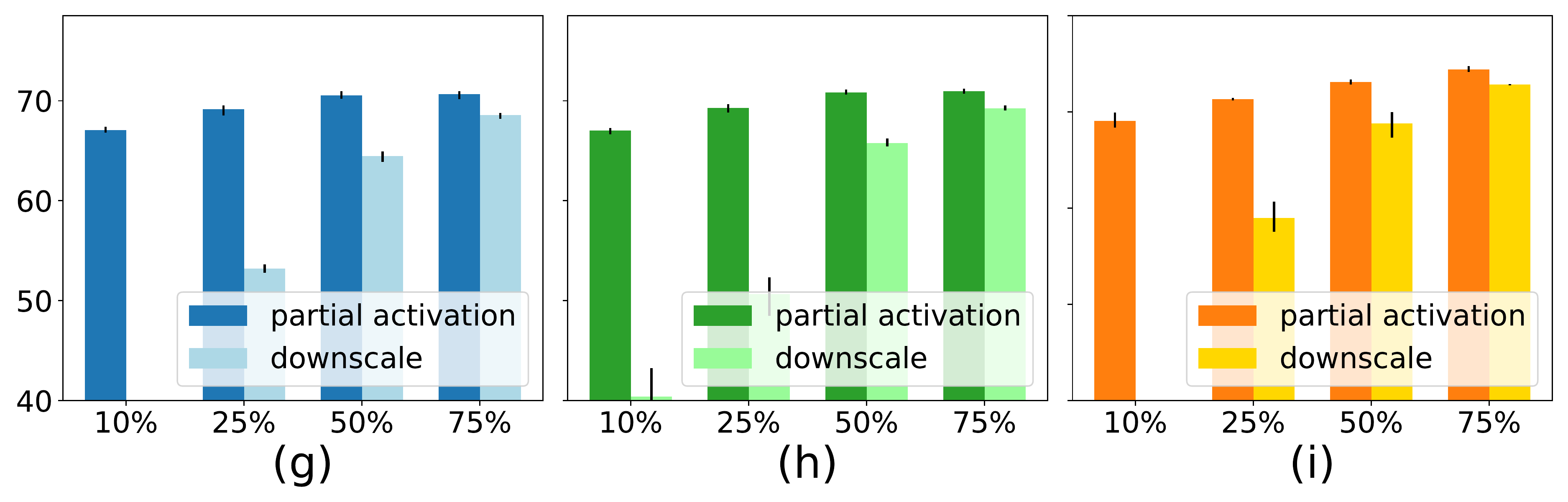}

\caption{Comparison between different partial activation ratios on SqueezeNet, ShuffleNetV2, and MobileNetV2, in terms of accuracy ((a) - CIFAR-10, (b) - CIFAR-100)  and communication complexity (c). Ratio of $50\%$ presents a good balance between accuracy and efficiency. (d)-(i): comparison between partial activation and reducing network width at the same ratios (i.e.\ removing the no-activation branch) between SqueezeNet ((d) - CIFAR-10, (g) - CIFAR-100), ShuffleNetV2 ((e) - CIFAR-10, (h) - CIFAR-100) and MobileNetV2 ((f) - CIFAR-10, (i) - CIFAR-100).  Results demonstrate that the channels with no ReLU activation still contributes significantly to the accuracy.}
\label{fig:partial_activation_acc_comm_and_downscale}
\vspace{-0.3cm}
\end{figure*}

In this section, we conduct a sequence of experiments demonstrating the effectiveness of our method in the design of crypto-oriented neural architectures. Our crypto-oriented architectures have better trade-offs between efficiency and accuracy in the privacy-preserving regime than standard architectures.

\paragraph{Efficiency evaluation metric.}
The fundamental complexity measures for secure computations are the communication and the round complexities, as they represent the structure of the interactive protocol. The runtime of a protocol is hardware and implementation specific, both having large variability. In this work we focused on the communication and round complexities, and provided the runtime only on the two extreme cases: removing all activation in Fig.~\ref{fig:noReluRuntime}, and using all our proposed optimizations in Table~\ref{tab:crypto_oriented_results}.

\paragraph{Implementation details.} We focused on the case of privacy-preserving inference and assumed the existence of trained models. For this reason, during experiments, we trained the different networks in the clear and “encrypted” them to measure accuracy, runtime, round complexity and communication complexity on private data. 
We use the tf-encrypted framework  \cite{tfencrypted} to convert trained neural networks to privacy-preserving neural networks. This implementation is based on secure multi-party computation and uses the SPDZ \cite{damgaard2012multiparty, damgaard2013practical} and SecureNN \cite{wagh2019securenn} protocols as backend. For runtime measurements we used an independent server for each party in the computation, each consisting of 30 CPUs and 20GB RAM. 

Due to limited resources, we evaluated on the CIFAR-10 and CIFAR-100 datasets \cite{krizhevsky2009learning}. Experiments were conducted on downscaled versions of three popular efficient architectures - SqueezeNet  \cite{iandola2016squeezenet}, ShuffleNetV2  \cite{ma2018shufflenet} and MobileNetV2  \cite{sandler2018mobilenetv2}. For more details on the downscaling, we refer the reader to the supplementary material.

\paragraph{Training details.} We train our models using stochastic gradient descent (SGD) optimizer and the Nestrov accelerated gradient and momentum of $0.9$. We use a cosine learning rate which starts from 0.1 (0.04 for SqueezeNet) and reduces to 0. In every experiment, we trained from scratch five times and report the average result.

%-------------------------------------------------------------------------

\subsection{Partial Activation Ratio} We experimented with different partial activation ratios between the channels in the non-linear branch and the total number of channels. Results are presented in Fig.~\ref{fig:partial_activation_acc_comm_and_downscale}. $50\%$ appears to be a good trade-off between efficiency and accuracy. Note that round complexity was eliminated from this comparison as we assume that element-wise non-linearities can all be computed in parallel, i.e.\ each round of interaction during the secure computation consists of the communication of all element-wise non-linearities in the layer. Under this reasonable assumption, the round complexity is constant across each layer regardless of the number of non-linearities in the layer.

\subsection{Scaling Down Network Width} The reduction of non-linearities across layers can also be achieved by simply scaling down the architecture's width, i.e.\ reducing the number of channels in each layer (equivalent to dropping the no-activation branch). We compared the performance of scaling down and using partial activation with the same ratio of remaining channels. As can be seen in Fig.~\ref{fig:partial_activation_acc_comm_and_downscale}, scaling down the width is inferior to the use of partial activation with the original width, demonstrating the importance of both branches in the partial activation layer.
Note that as we enlarge the non-linear branch in the partial activation layer or reduce the number of removed channels in the downscaling, the difference to the original model decrease, resulting with a reduction in the accuracy loss. 

%-------------------------------------------------------------------------

\subsection{Activation Removal}  We evaluated the effectiveness of removing activation layers from each of the three architecture blocks, where each block has two activation layers. Our experiment exhaustively evaluates the effects of removing one or both layers. The results presented in Table~\ref{tab:activation_removal} clearly demonstrate that one activation layer in each block can be removed completely with a plausible loss of accuracy.

\subsection{Activation Removal and Partial Activation} In order to further minimize the use of non-linearities, we investigated the combination of our \textit{partial activation} layer with complete removal of other activation layers. We evaluated the removal of one activation layer while replacing the other with a $50\%$-partial activation layer. Results are presented in Table~\ref{tab:activation_removal} and demonstrate that we were able to significantly improve the communication complexity of the secure inference of SqueezeNet, ShuffleNetV2 and MobileNetV2 by $26.3\%$, $49.4\%$ and $63.3\%$, respectively, with a minor change in accuracy. The round complexity was considerably improved as well with $20.4\%$, $39.2\%$ and $42.2\%$ improvement for SqueezeNet, ShuffleNetV2 and MobileNetV2, respectively.

\begin{table}[tb]
\centering
\small
    \begin{tabular}{lccc}
    \toprule
    \textbf{Model} & \textbf{Accuracy} &\textbf{Comm. } &\textbf{Rounds}\\
    & {\scriptsize CIFAR-10 /\ 100} & (MB) & \\
    \midrule
    Squeeze-1 & 90.54 /\ 64.72 & 189.36 & 233 \\
    Squeeze-2 & 93.15 /\ 72.37 & 309.97 & 313 \\
    Squeeze-0.5-1 & 90.4 /\ 66.04 & 180.75 & 233 \\
    Squeeze-0.5-2 & 92.66 /\ 70.76 & 241.05 & 313 \\
    Squeeze-none & 86.95 /\ 60.03 & 172.13 & 153 \\
    Squeeze-orig & 92.49 /\ 70.41 & 327.2 & 393 \\
    \midrule
    Shuffle-1 & 92.83 /\ 71.02 & 219.01 & 294 \\
    Shuffle-2 & 92.9 /\ 71.37 & 188.86 & 324 \\
    Shuffle-0.5-1 & 92.5 /\ 70.07 & 157.63 & 294 \\
    Shuffle-0.5-2 & 92.19 /\ 69.53 & 142.55 & 324 \\
    Shuffle-none & 83.26 /\ 46.8 & 95.25 & 134 \\
    Shuffle-orig & 92.6 /\ 70.95 & 311.63 & 484 \\
    \midrule
    Mobile-1 & 93.92 /\ 74.35& 1168.2 & 466 \\
    Mobile-2 & 94.13 /\ 74.17 & 1003.02 & 486 \\
    Mobile-0.5-1 & 93.66 /\ 72.77 & 706.54 & 466 \\
    Mobile-0.5-2 & 93.28 /\ 72.86 & 623.95 & 486 \\
    Mobile-none & 78.45 /\ 51.45 & 244.88 & 146 \\
    Mobile-orig & 94.49 /\ 74.8 & 1926.34 & 806 \\
	\bottomrule
    \end{tabular}

\caption{Effects of the removal of activations in network blocks. Tested networks are built of blocks having two activation layers each.
In $Network$-$i$ we keep only the $i'th$ activation layer in each block, and remove the other activation layer.
In $Network$-0.5-$i$ we replace the $i'th$ activation layer with a $50\%$ partial activation layer, and remove the other activation layer.
We also show the original blocks and the removal of all activation layers in each block.
The table shows that removing most of the activations in each block has minimal effect on accuracy but substantially increase speed.
}

\label{tab:activation_removal}
% \vspace{-0.5cm}
\end{table}
%-------------------------------------------------------------------------
\subsection{Alternative Non-Linearities}
\label{sec:other_nonlinear}

Secure computation of non-linear layers is costly, but the cost of different non-linearities varies significantly. In addition to the removal of non-linearities, we investigated the cost of several commonly used non-linearities and propose more crypto-oriented alternatives.

\begin{figure}[tb]
\centering
\includegraphics[width=0.99\linewidth,height=0.9\linewidth]{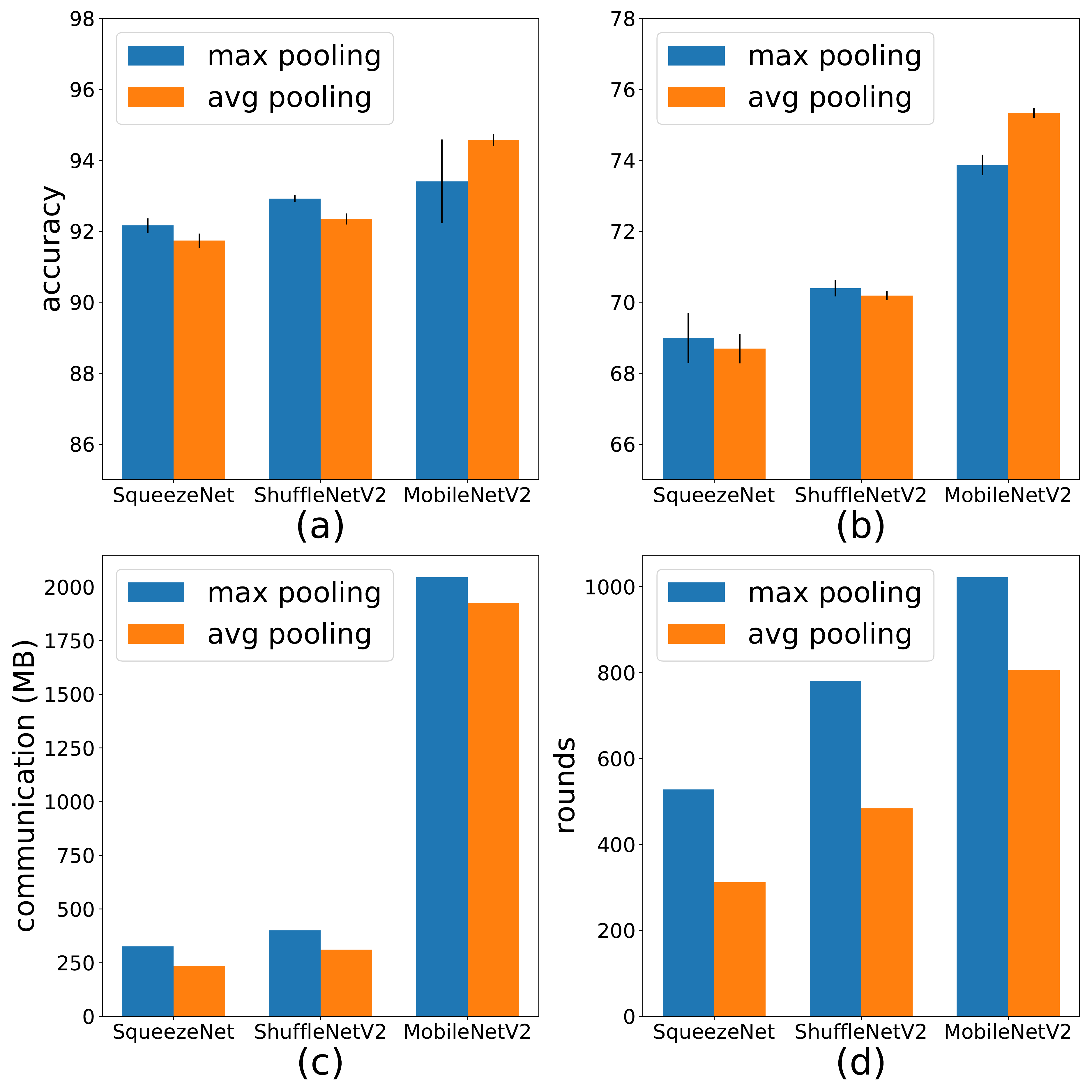}

\caption{Comparison between max pooling (blue) and average pooling (orange) on SqueezeNet, ShuffleNetV2 and MobileNetv2, in terms of accuracy ((a) - CIFAR-10, (b) - CIFAR-100), communication complexity (c) and round complexity (d). Average pooling has similar accuracy but is much more efficient.}
\label{fig:pooling}
\vspace{-0.3cm}
\end{figure}

\begin{figure}[tb]
\centering
\includegraphics[width=0.99\linewidth,height=0.9\linewidth]{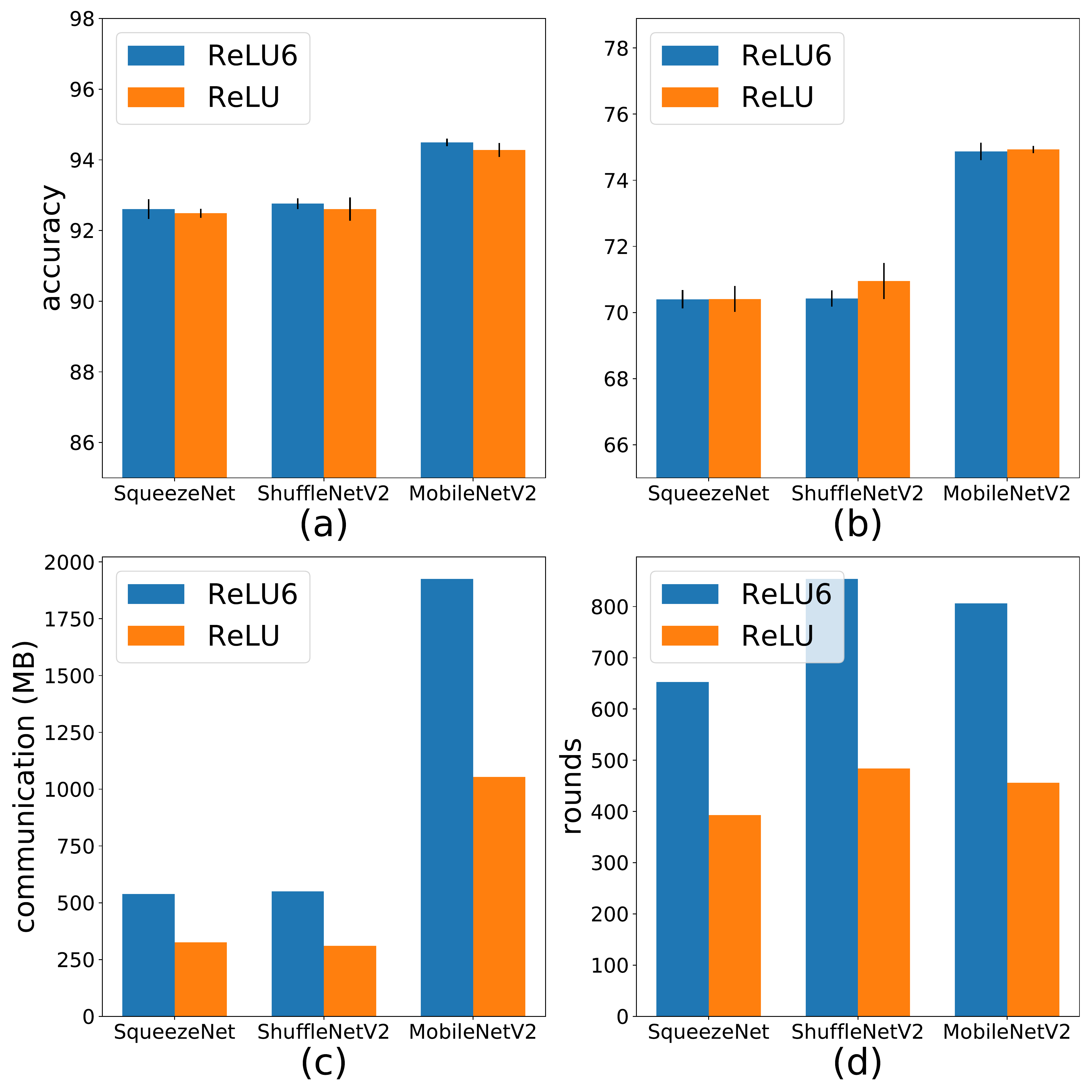}

\caption{Comparison between ReLU6 (blue) and ReLU (orange) activation functions on SqueezeNet, ShuffleNetV2 and MobileNetv2, in terms of accuracy ((a) - CIFAR-10, (b) - CIFAR-100), communication complexity (c) and round complexity (d). Accuracy is similar but ReLU is more efficient that ReLU6.}
\label{fig:relu6}
\vspace{-0.3cm}
\end{figure}

\paragraph{Pooling.} 
Previous empirical results show that replacing max pooling with average pooling has minimal effect on network accuracy and (non-secure) inference runtime. Many recent neural networks use both pooling methods, or replace some of them with strided convolutions, which are a computationally efficient approach to average pooling \cite{ioffe2015batch, he2016deep, szegedy2016rethinking, szegedy2017inception, chollet2017xception, huang2017densely, sandler2018mobilenetv2, tan2019efficientnet}. 
In secure inference of neural networks, max and average pooling have very different costs. While max pooling is a non-linear operation which requires computing a complicated protocol, average pooling can be simply performed by summation and multiplication with a constant scalar. For example, in the SecureNN  \cite{wagh2019securenn} protocol, the max pooling layer has a round complexity of:
\begin{equation}
\label{eq:maxpool_rounds}
 Rounds_{MaxPooling} = 9(f^2-1)   
\vspace{-0.09cm}
\end{equation}
 here $f\times f$ is the kernel area, and a communication complexity of 
 \begin{equation}
 \label{eq:maxpool_comm}
 Comm_{MaxPooling} = (8l\log p +29l)(f^2-1)
 \vspace{-0.09cm}
 \end{equation}
 where $l$ is the number of bits representing the input numbers and $p$ denotes the field size - each $l$-bit number is secret shared as a vector of $l$ shares, each being a value between 0 and $p-1$ (inclusive).
Consider a pooling layer with input image of size $32 \times 32 \times 3$ and pooling kernel size of $2 \times 2$. Max pooling would require $27$ rounds and $1.43$ MB communication, whereas average pooling can be computed locally by each party, i.e.\ with no communication required. 

We evaluated the effect of using max pooling against average pooling. SqueezeNet consists of multiple max pooling layers and a global average pooling layer. In the max pooling experiment, we replaced the global pooling layer with a max global pooling, while in the average pooling experimented we replaced all max pooling layers with average pooling. MobileNetV2 and ShuffleNetv2 use strided convolutions for dimensionality reduction. In order to better emphasize the effect of the different pooling methods, we removed the strides and replaced them with pooling layers. Results are presented in Fig.~\ref{fig:pooling}. We can see that average pooling is much more efficient while not affecting accuracy significantly in comparison to max pooling.

\paragraph{ReLU6.} 
Many variants were proposed for the ReLU activation function with the objective of improving the training procedure. One common variant is the ReLU6 activation  \cite{krizhevsky2010convolutional}, which is defined as:
\begin{equation}
\label{eq:relu}
ReLU6(x) = min(max(x, 0), 6)
 \vspace{-0.09cm}
\end{equation} 
This activation function is used in several recent efficient architectures including MobileNetV2  \cite{sandler2018mobilenetv2}. As mentioned in Sec.~\ref{sec:background}, comparisons are difficult to compute in a secure manner. Therefore, the cost of ReLU6, which consists of two comparisons is double the cost of the standard ReLU activation. We provide a protocol for the secure computation of ReLU6 and corresponding analysis in the supplementary material. 

We investigated the effect of using ReLU6 versus ReLU activations. MobileNetV2 was designed with ReLU6 so we simply replace those with ReLU. ShuffleNetV2 and SqueezeNet use the ReLU activation, which we replaced with the ReLU6 activation. Results are presented in Fig.~\ref{fig:relu6}. The choice of non-linearity has minimal effect on accuracy, while ReLU is twice as efficient as ReLU6. 

%-------------------------------------------------------------------------
\subsection{Crypto-Oriented Neural Architectures}
\label{sec:final_exp}
\begin{figure}[tb]
\centering
\includegraphics[width=0.99\linewidth, height=0.48\linewidth]{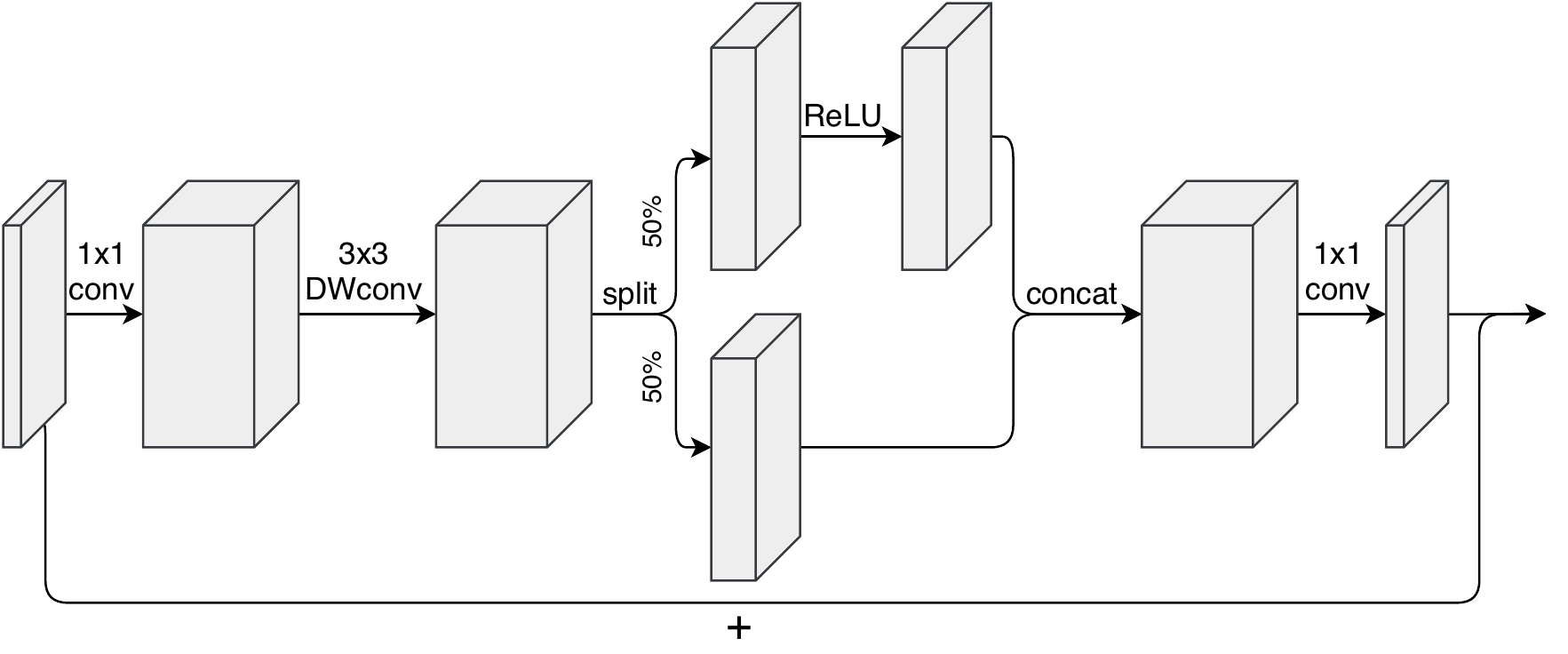}
\caption{Crypto-oriented inverted residual with linear bottleneck block. By applying our method, i.e.\ removing the first activation layer, replacing the second activation layer with $50\%$-partial activation and using the ReLU activation instead of the ReLU6 variant, we achieved a significant improvement in communication and round complexity of the MobileNetV2 architecture.}
\label{fig:our_mbconv}
\vspace{-0.3cm}
\end{figure}

We use our method to design state-of-the-art crypto-oriented neural network architectures, based on regular state-of-the-art architectures. Specifically, we present crypto-oriented versions of the building blocks in SqueezeNet  \cite{iandola2016squeezenet}, ShuffleNetV2  \cite{ma2018shufflenet} and MobileNetV2  \cite{sandler2018mobilenetv2}. For illustrative purposes, we describe in detail the application of our method on the inverted residual with linear bottleneck blocks from MobileNetV2 with the CIFAR-10 dataset, illustrated in Fig.~\ref{fig:our_mbconv}. A more detailed description of the applications of our method for the other $2$ blocks and other datasets is presented in the supplementary material. Final results on CIFAR-10, CIFAR-100, MNIST \cite{lecun1998gradient} and Fashion-MNIST \cite{xiao2017fashion} are presented in Table~\ref{tab:crypto_oriented_results}. 

In order to reduce the number of non-linear evaluations we replaced all activation layers in the inverted residual with linear bottleneck block with $50\%$-partial activation layers. This results with an improvement of $43.6\%$ in communication complexity.

After careful evaluation, we removed the first activation layer completely (i.e.\ the depthwise convolution is the only non-linear layer). This reduces the communication complexity by $47.9\%$ and the round complexity by $39.8\%$. Combining this change with the former, i.e.\ replacing the second activation layer by a $50\%$ partial activation results with additional improvement of $37.8\%$ in communication complexity. Overall, by applying both changes we reduce the communication and round complexity by $67.6\%$ and $39.8\%$, respectively. 

As discussed in Sec.~\ref{sec:other_nonlinear}, the ReLU6 variant costs twice as much as the ReLU activation. Therefore, we replace the ReLU6 activation function with the ReLU function. This change produces an improvement of $45.3\%$ in communication complexity and $43.6\%$ in round complexity. 

The above modifications yield an improvement of $79\%$ in communication complexity and $63.4\%$ in round complexity. 

%-------------------------------------------------------------------------
\section{Discussion}
\label{sec:discussion}

\paragraph{Accuracy measurements.} Due to the slow inference runtime of secure neural networks, we have measured accuracy in the non-secure setting. As the SecureNN framework, which we base our analysis on, applies no approximations on the inference, there should be no significant difference between the secure and non-secure accuracy. In order to verify this assumption, we have measured the secure inference accuracy on a subset of experiments, using the tf-encrypted library \cite{tfencrypted}, and compared to the non-secure accuracy of the same model. The results, presented in the supplementary material, show minor loss of accuracy.

\paragraph{Comparison to other frameworks.} Our analysis focused on the SecureNN protocol \cite{wagh2019securenn}. We stress that the same idea applies to other frameworks. For example, consider the Gazelle framework \cite{juvekar2018gazelle}, which proposed a hybrid between homomorphic encryption for linear layers, and garbled circuits for non-linear layers. According to the benchmarks provided by the authors, a convolution layer with input of size $32\times 32 \times 32$, kernel size $3$ and $32$ output channels, i.e.\ $32,768$ output neurons, will take $899$ms and no communication. In comparison, a ReLU layer with $10,000$ neurons, less than third of the output of the aforementioned convolution layer, will take $1,858$ms and $71.1$MB.

It should be noted that there are frameworks for which secure computation of a convolution layer is more expensive then secure computation of a ReLU layer. An example is the DeepSecure framework \cite{rouhani2018deepsecure} which only uses garbled circuits. The optimal architectures might change slightly in this case, but the core idea of our work, i.e.\ the need for designing crypto-oriented architectures, is highly relevant.

\paragraph{Increasing channels.} In order to minimize accuracy reduction, we tried to gain more expressiveness by increasing the number of channels with no activation.
As discussed in Sec.~\ref{sec:method}, based on the analysis of the SecureNN \cite{wagh2019securenn} framework (and others, as discussed above), secure computation of convolution layers is more efficient than the cost of activation layers. Therefore we can add more channels when removing non-linearities. Results are detailed in the supplementary material and show a minor increase in accuracy while slightly increasing communication. The difference was not significant enough to be included in our final crypto-oriented architectures.

%-------------------------------------------------------------------------
\section{Conclusion}
We addressed efficiency challenges in privacy-preserving neural network inference. Motivated by the unique properties of secure computation, we proposed a novel activation layer for crypto-oriented neural network architectures: partial activation layers. By using our activation layer, together with selective removal of some activation layers and avoiding the use of expensive non-linear variants, on three state-of-the-art architectures (SqueezeNet, ShuffleNetV2 and MobileNetV2) and various datasets (CIFAR-10, CIFAR-100, MNIST and Fashion-MNIST) we achieved significant improvement on all architectures. For MobileNetV2 on the CIFAR-10 dataset, for example, we achieved an improvement of $79\%$ in communication complexity, $63\%$ in round complexity and $58\%$ in secure inference runtime, with only a reasonable loss in accuracy.
%------------------------------------------------------------------------

%------------------------------------------------------------------------
\subsubsection*{Acknowledgments}
This research has been supported by the Israel ministry of Science and Technology, by the Israel Science foundation, and by the European Union's Horizon 2020 Framework Program (H2020) via an ERC Grant (Grant No.\ 714253).
%------------------------------------------------------------------------

{\small
\bibliographystyle{ieee_fullname}
\bibliography{main}
}

%------------------------------------------------------------------------
\renewcommand{\thesection}{\Alph{section}}
\renewcommand{\thesubsection}{A.\arabic{subsection}}
\section*{Appendix}
%------------------------------------------------------------------------

\subsection{Architecture Downscaling}
Experiments were conducted on three popular efficient architectures -  SqueezeNet~\cite{iandola2016squeezenet}, ShuffleNetV2~\cite{ma2018shufflenet} and MobileNetV2~\cite{sandler2018mobilenetv2}. These architectures were designed for the ImageNet dataset~\cite{deng2009imagenet}, a large scale dataset. Due to limited resources we evaluated on smaller datasets - CIFAR-10, CIFAR-100, MNIST and Fashion-MNIST \cite{krizhevsky2009learning, lecun1998gradient, xiao2017fashion} which required downscaling the architectures accordingly. We will detail the modifications applied to each architecture:

\paragraph{SqueezeNet.} Changed the kernel size of the first convolution layer from $7\times 7$ to $3\times 3$ and reduced the stride from $2$ to $1$. In each pooling layer, we replaced the $3\times 3$ kernel with a $2\times 2$ kernel. In addition, we added batch normalization layers after every convolution layer. In CIFAR-10, CIFAR-100 and Fashion-MNIST we removed the dropout layer while in MNIST we added a $0.1$ dropout layer at the end of each block.

\paragraph{ShuffleNetV2.} Reduced the stride in the first convolution layer from $2$ to $1$. In addition, we removed the first pooling layer. 

\paragraph{MobileNetV2.} Reduced the stride in the first convolution layer and in the first inverted residual block from $2$ to $1$. In addition, we increased the weight decay from $0.00004$ to $0.0002$. For CIFAR-100 and Fashion-MNIST we have used a dropout rate of $0.1$.

%-------------------------------------------------------------------------
\subsection{Other Crypto-Oriented Neural Architectures}
We present our crypto-oriented version of the building blocks in SqueezeNet~\cite{iandola2016squeezenet} and ShuffleNetV2~\cite{ma2018shufflenet}. 
\subsubsection{SqueezeNet}
\begin{figure}[tb]
\centering
%w49 h35
\includegraphics[width=0.99\linewidth, , height=0.7\linewidth]{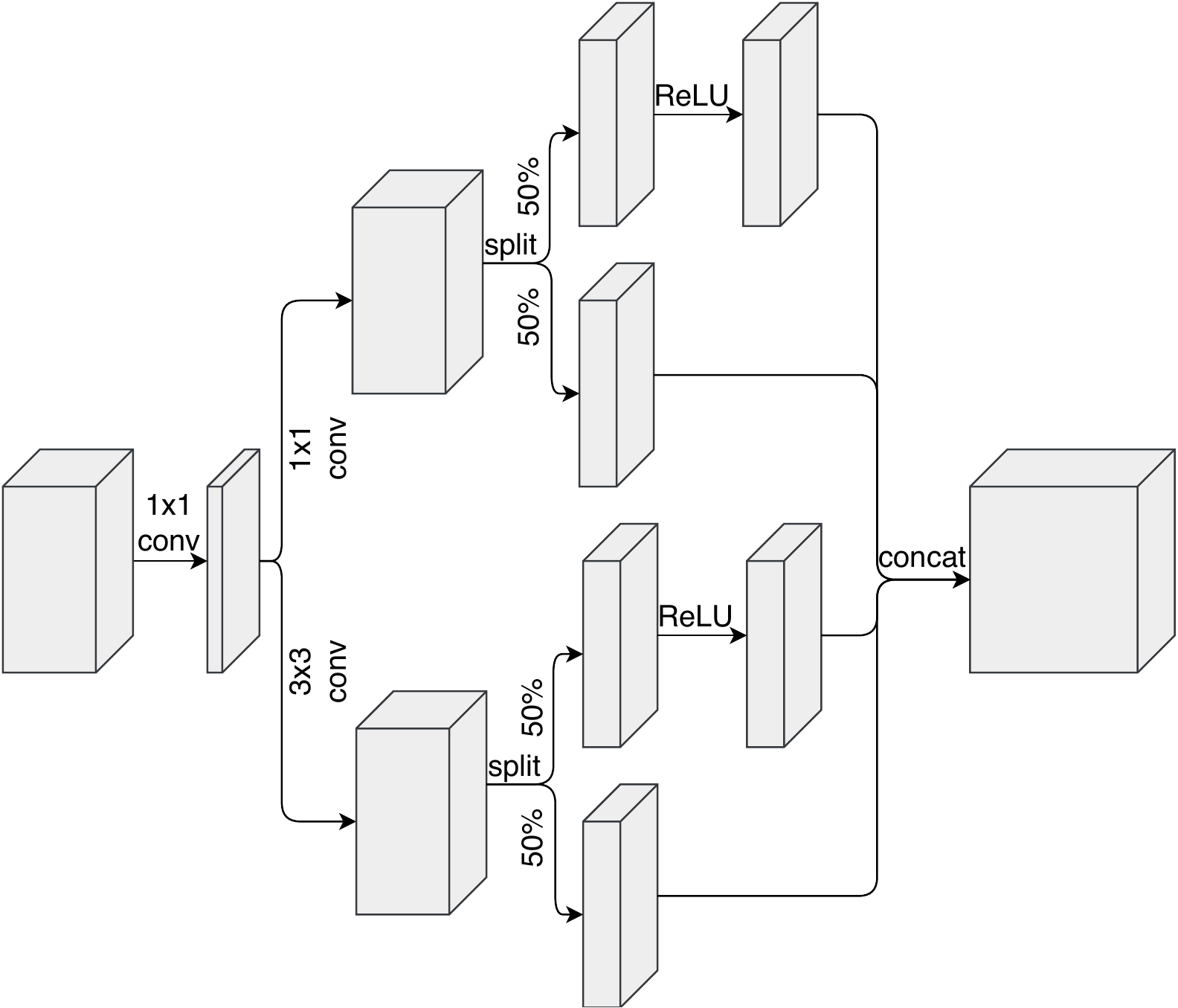}
\caption{Crypto-oriented Fire module block. By applying our design method, i.e.\ removing first activation layer and replacing second activation layer with a partial activation layer, we achieved a significant improvement in communication and round complexity of the SqueezeNet architecture, with a reasonable accuracy loss.}
\label{fig:our_squeezenet}
\vspace{-0.3cm}
\end{figure}

In order to reduce the number of non-linearities we replaced the activation layers in the \textit{Fire module} (i.e.\ the SqueezeNet building block) with partial activation layers, and used a ratio of $50\%$ for the CIFAR-10 and CIFAR-100 datasets and $25\%$ for the MNIST and Fashion-MNIST datasets. This reduces the communication complexity by $23.7\%$ for the two CIFAR datasets and $35.56\%$ for the two MNIST datasets. 

For the two CIFAR datasets we removed the first activation layer in the \textit{Fire module}, i.e.\ from the squeeze phase of the block. This results with an improvement of $5.26\%$ in communication complexity and with $20.36\%$ in round complexity. For the two MNIST datasets we removed the second activation layer, which results with $42.15\%$ and $40.71\%$ improvement in communication and round complexity, respectively. 
Combining this change with the former, i.e.\ replacing the remaining activations with partial activation layers results further improves the communication complexity by $22.23\%$ and $19.31\%$ for the two CIFAR and the two MNIST, respectively.

Overall, by applying both changes we reduce communication and round complexity by $26.39\%$ and $20.36\%$, respectively, for CIFAR-10 and CIFAR-100. For MNIST and Fashion-MNIST we reduce communication complexity by $46.1\%$ and round complexity by $40.71\%$.

We replaced each max pooling layer with an average pooling layer, as max pooling is very expensive to compute in a secure manner. This improved the communication complexity by $28.02\%$ for CIFAR and $26.97\%$ for MNIST, and the round complexity by $20.61\%$. 
\\\\
Our final crypto-oriented version of the SqueezeNet architecture is improving over it's non crypto-oriented counterpart by $54.41\%$ in communication complexity and $40.97\%$ in round complexity, for CIFAR-10 and CIFAR-100, at the cost of a reasonable accuracy loss of $0.6\%$ and $0.7\%$, respectively. Our CIFAR crypto-oriented \textit{Fire module} block is presented in Fig.~\ref{fig:our_squeezenet}. For MNIST and Fashion-MNIST we improve communication complexity by $73.08\%$ and round complexity by $61.32\%$ with accuracy loss of $0.19\%$ for MNIST and $0.7\%$ for Fashion-MNIST.

\subsubsection{ShuffleNetV2}

\begin{figure}[tb]
\centering
\includegraphics[width=1.\linewidth, height=0.5\linewidth]{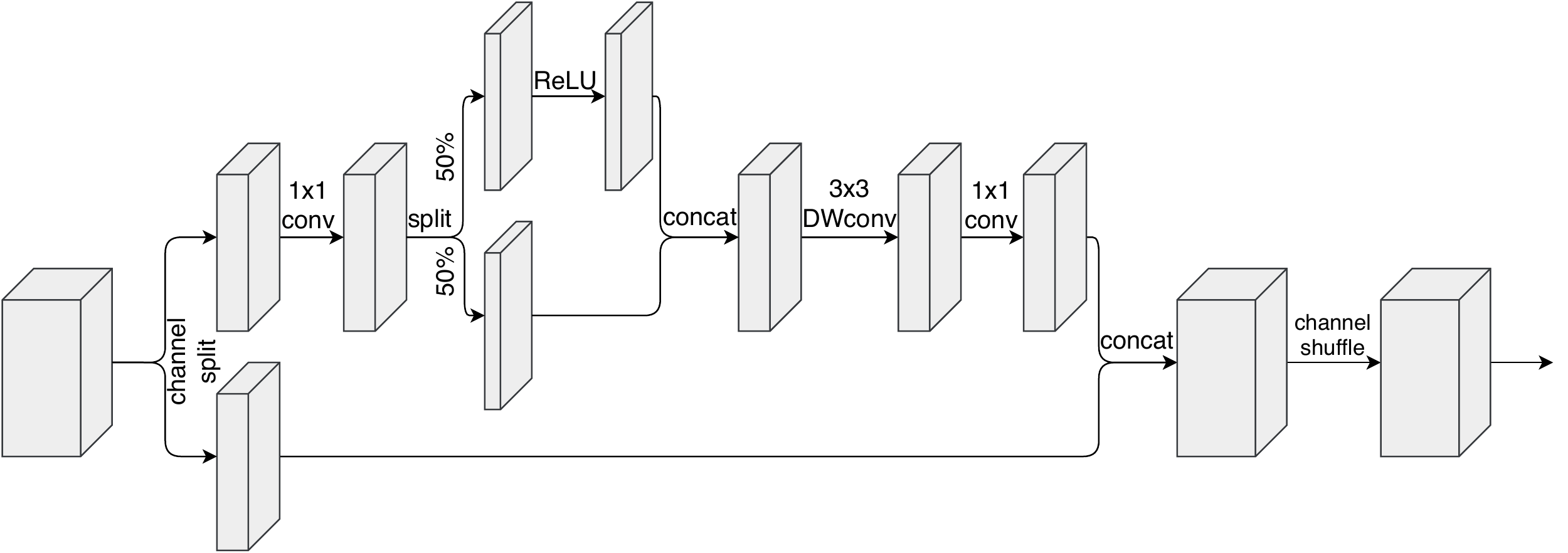}
\caption{Crypto-oriented ShuffleNetV2 unit. By applying our method, i.e.\ removing second activation layer and replacing first activation layer with partial activation, we achieved a significant improvement in communication and round complexity of the ShuffleNetV2 architecture with only a small loss of accuracy.}
\label{fig:our_shufflenet}
\vspace{-0.3cm}
\end{figure}

We replaced all activation layers in the ShuffleNetV2 unit by $50\%$-partial activation layers, for CIFAR-10 and CIFAR-100, and $25\%$-partial activation layers for MNIST and Fashion-MNIST. This improves the communication complexity by $34.56\%$ and $50.81\%$, respectively.

 We removed the second activation layer from the ShuffleNetV2 unit. This results with an improvement of $29.71\%$ in communication complexity and $39.26\%$ in round complexity, for CIFAR-10 and CIFAR-100. For MNIST and Fashion-MNIST this results with an improvement of $29.44\%$ in communication complexity.
 
 By further reducing the number of non-linearities and replacing the remaining activation layer with a partial activation layer we get an additional improvement of $28.02\%$ and $40.71\%$ in communication complexity for the two CIFAR datasets and two MNIST datasets, respectively. 

We did not change the non-linearities in the architecture as the ShuffleNetV2 unit was not using expensive variants such as ReLU6. In addition, the max pooling layer that existed in the original design of the architecture was removed in our downscaling process, detailed above. 
\\\\
Applying the aforementioned modifications yields an improvement of communication and round complexity by $49.41\%$ and $39.26\%$, respectively, for the CIFAR-10 and CIFAR-100 datasets. This optimization has a reasonable accuracy loss of $0.1\%$ for CIFAR-10, and $0.88\%$ for CIFAR-100. Our CIFAR crypto-oriented ShuffleNetV2 unit is presented in Fig.~\ref{fig:our_shufflenet}. For MNIST and Fashion-MNIST, our modifications yields an improvement of $58.17\%$ in communication complexity with a small accuracy loss of $0.03\%$ and $0.11\%$ for MNIST and Fashion-MNIST, respectively.

\subsubsection{MobileNetV2}
As detailed in the experiments section, for the CIFAR-10 and CIFAR-100 datasets we have replaced all ReLU6 activations in the inverted residual with linear bottleneck block by ReLU activations. In addition, we removed the first activation layer in the block completely and replaced the second with a $50\%$-partial activation layer. Overall, these modifications yield an improvement of $79\%$ in communication complexity and $63.4\%$ in round complexity, with a reasonable accuracy loss of $1.05\%$ for CIFAR-10 and $2.19\%$ for CIFAR-100. 

For the MNIST and Fashion-MNIST we have applied the same modification, with the only difference in the partial activation ratio - $25\%$ rather than $50\%$. Overall, the modifications result with an improvement of $83.5\%$ in communication complexity, with small accuracy loss of $0.22\%$ for Fashion-MNIST and a minor accuracy gain of $0.02\%$ for MNIST. 

%-------------------------------------------------------------------------
\subsection{Encrypted Accuracy}

\begin{table}[tb]
\centering
\small
    \begin{tabular}{lcc}
    \toprule
    \textbf{Model} & \textbf{Secure} &\textbf{Non-Secure}\\
     & \textbf{Accuracy} & \textbf{Accuracy}\\
    \midrule
    CO-SqueezeNet & 91.88 & 91.89 \\
    CO-ShuffleNetV2 & 92.49 & 92.51 \\
    CO-MobileNetV2 & 93.46 & 93.44\\
	\bottomrule
    \end{tabular}
    \vspace{.2cm}
%\end{center}
\caption{Comparison of secure and non-secure CIFAR-10 accuracies on our crypto-oriented architectures. We denote by \textit{CO-Network} our crypto-oriented variant. There is a negligible difference in accuracy between the secure and non-secure setting.}

\label{tab:enc_accuracy}
\vspace{-0.3cm}
\end{table}

In our experiments, we measured accuracy in the non-secure setting, due to the slow inference time of secure neural networks. Our experiments were conducted using the tf-encrypted library~\cite{tfencrypted} which is based on the SPDZ~\cite{damgaard2012multiparty, damgaard2013practical} and SecureNN~\cite{wagh2019securenn} protocols. As this implementation does not apply any approximations on the inference, we do not expect there to be a significant difference between the secure and non-secure accuracy measurements. In order to verify this assumption, we evaluated the secure accuracy on our final crypto-oriented architectures and compared the results against the non-secure accuracy, on the CIFAR-10 dataset. As done in all of our experiments, each experiment was conducted five times, and we report the average results. The results presented in Table~\ref{tab:enc_accuracy} show that the accuracy difference is indeed negligible.

%-------------------------------------------------------------------------
\subsection{Double Channels Results}

\begin{table}[tb]
\centering
\small
    \begin{tabular}{lccc}
    \toprule
    \textbf{Model} & \textbf{Accuracy} &\textbf{Comm. (MB)} &\textbf{Rounds}\\
    \midrule
    Sq-$0.5\cdot 1$\ts{st} & 90.4 & 179.95 & 233 \\
    Sq-$0.5\cdot 1$\ts{st}-double & 90.66 & 241.75 & 233 \\
    Sq-$0.5\cdot 2$\ts{nd} & 92.66 & 240.26 & 313 \\
    Sq-$0.5\cdot 2$\ts{nd}-double & 92.98 & 256.74 & 313\\
    Sq-orig & 92.49 & 326.41 & 393 \\
    \midrule
    Sh-$0.5\cdot 1$\ts{st} & 92.5 & 156.89 & 294 \\
    Sh-$0.5\cdot 1$\ts{st}-double & 92.81 & 308.41 & 294 \\
    Sh-$0.5\cdot 2$\ts{nd} & 92.19 & 141.82 & 324 \\
    Sh-$0.5\cdot 2$\ts{nd}-double & 92.46 & 188.84 & 324 \\
    Sh-orig & 92.6 & 310.89 & 484 \\
    \midrule
    Mb-$0.5\cdot 1$\ts{st} & 93.66 & 705.62 & 466 \\
    Mb-$0.5\cdot 1$\ts{st}-double & 94.15 & 1397.7 & 466 \\
    Mb-$0.5\cdot 2$\ts{nd} & 93.28 & 623.03 & 486 \\
    Mb-$0.5\cdot 2$\ts{nd}-double & 94.15 & 1213.51 & 486\\
    Mb-orig & 94.49 & 1925.42 & 806 \\
	\bottomrule
    \end{tabular}
    \vspace{.2cm}
%\end{center}
\caption{Comparison of the effect of increasing the number of channels in convolution layers without activations. In this experiments, we removed different activation layers (none, 1st, 2nd or both) and replaced the remaining with a $50\%$ partial activation layer. This was performed on SqueezeNet (\textit{Sq}), ShuffleNetV2 (\textit{Sh}) and MobileNetV2 (\textit{Mb}) blocks. By \textit{Sq-$0.5\cdot i$-double
}, we denote: i) the removal of all but the $i\ts{th}$ activation layer ii) its replacement with a $50\%$ partial activation layer iii) doubling the amount of channels in the no-activation convolution layer. Results show minor minor effect on CIFAR-10 accuracy while increasing the communication complexity.}

\label{tab:double_channel_results}
\vspace{-0.3cm}
\end{table}

As mentioned in the discussion section, we tried to reduce the accuracy loss resulting from the minimization of activations, i.e.\ the removal of activation layers and replacement of the remaining layers with partial activation layer, described in Table~\ref{tab:double_channel_results}. We evaluated the effect of increasing the number of channels in layers with no activations by a factor of two. The goal is to amplify the model's expressiveness, without adding further non-linearities. As mentioned in the paper and based on the analysis of the SecureNN framework~\cite{wagh2019securenn}, the added cost of increasing the convolution channels is less then the cost of the removed activations, therefore enabling us to "compensate" for the removal of activations with more convolutional channels. As can be in the results, detailed in Table~\ref{tab:double_channel_results}, increasing the amount of channels had a minor affect on CIFAR-10 accuracy while increasing the costs. The benefit of increasing the channels was not significant enough to be included in our final crypto-oriented architectures.

%-------------------------------------------------------------------------

\subsection{LeakyReLU Protocol}
We present a protocol, based on the ReLU protocol from ~\cite{wagh2019securenn}, for the secure computation of the LeakyReLU activation function. The LeakyReLU activation is defined as:
\begin{equation}
\label{eq:leakyReLUDef}
LeakyReLU(x) = max(0.1x, x)
\vspace{-0.09cm}
\end{equation} 
This activation function can be also written as:
\begin{equation}
\label{eq:leakyReLU}
LeakyReLU(x) = x(0.1 + 0.9\cdot H(x))
\vspace{-0.09cm}
\end{equation} 

Note that the secure computation of LeakyReLU only differs from the secure computation of ReLU, provided in ~\cite{wagh2019securenn}, by only a constant scalar multiplication, and therefore has the same communication and round complexity. This suggests that the LeakyReLU activation function can be used instead of ReLU with no additional costs.

\begin{mdframed}
    \textbf{Algorithm 2.} $\prod_{\sf LeakyReLU}(\{P_0,P_1\},P_2)$:\\
    \textbf{Input:} $P_0,P_1$ hold $\langle a \rangle _0 ^L$ and $\langle a \rangle _1 ^L$, respectively. \\
    \textbf{Output:} $P_0,P_1$ get $\langle$LeakyReLU$(a)\rangle _0 ^L$ and $\langle$LeakyReLU$(a)\rangle _1 ^L$.\\
    \textbf{Common Randomness:} $P_0,P_1$ hold random shares of 0 over $\mathbb{Z}_L$,
    denoted by $u_0$ and $u_1$ resp. \\
    \parbox{\textwidth}{\begin{enumerate}
        \item $P_0, P_1, P_2$ run $\prod _{\sf DReLU}(\{P_0, P_1\},P_2)$ with $P_j,j\in\{0,1\}$ having input $\langle a \rangle _j ^L$ and $P_0, P_1$ learn $\langle \alpha \rangle _0 ^L$ and $\langle \alpha \rangle _1 ^L$, resp.
        
        \item $P_0, P_1, P_2$ call $\prod _{\sf MatMul}(\{P_0, P_1\},P_2)$ with  $P_j,j\in\{0,1\}$ having input $(\langle a \rangle _j ^L, \langle 0.1 + 0.9\alpha \rangle _j ^L)$ and $P_0, P_1$ learn $\langle c \rangle _0 ^L$ and $\langle c \rangle _1 ^L$, resp.
        
        \item For $j \in \{0,1\}$, $P_j$ outputs $\langle c\rangle _j ^L + u_j$.
    \end{enumerate}}%
\vspace{-0.5cm}
\label{alg:leaky_algo}
\end{mdframed}
%-------------------------------------------------------------------------

\subsection{ReLU6 Protocol}
We provide a protocol for the secure computation of ReLU6, based on the the ReLU protocol from ~\cite{wagh2019securenn}. The protocol is a step-by-step secure evaluation of the ReLU6 function, when decomposed into a combination of Heaviside step functions ($H$):
\vspace{-0.09cm}
\begin{gather}
ReLU6(x) = H(x)\cdot(x + (6-x)\cdot H(x-6))
\label{eq:relu6}\\
H(x) = 
\begin{cases}
    1, & \text{if } x\geq 0\\
    0, & \text{otherwise}
\end{cases}
\label{eq:heaviside}
\end{gather}

We denote the model provider by $P_0$ and the data owner by $P_1$. $P_2$ represents the \textit{crypto-producer}, a third-party ``assistant''  that provides randomness. $\langle a \rangle _0 ^L$ and $\langle a \rangle _1 ^L$ are the two secret shares of $a$ over $\mathbb{Z}_L$. $\prod_{\sf DReLU}$ and $\prod_{\sf MatMul}$ are the secure protocols presented in \cite{wagh2019securenn} for computing $H$ and matrix multiplication (for scalar multiplication, we use MatMul with $1\times 1$ matrices), respectively. For more details we refer the reader to \cite{wagh2019securenn}. The proof of this protocol follows directly via the security of the two underlying protocols and the fact that at each point in time the parties learn only secret shares of the current state of the computation.

The round and communication complexities of ReLU6 under this protocol are specified in Eq.~\eqref{eq:relu6_rounds}--\eqref{eq:relu6_comm}.
\begin{gather}
Rounds_{ReLU6} = 20
\label{eq:relu6_rounds}\\
Comm_{ReLU6} = 16l\log p + 48l
\label{eq:relu6_comm}
\end{gather}

The ReLU6 protocol involves two secure evaluations of the Heaviside step function and therefore requires twice the cost of ReLU. 

\begin{mdframed}
    \textbf{Algorithm 1} $\prod_{\sf ReLU6}(\{P_0,P_1\},P_2)$:\\
    \textbf{Input:} $P_0,P_1$ hold $\langle a \rangle _0 ^L$ and $\langle a \rangle _1 ^L$, respectively. \\
    \textbf{Output:} $P_0,P_1$ get $\langle$ReLU6$(a)\rangle _0 ^L$ and $\langle$ReLU6$(a)\rangle _1 ^L$. \\
    \textbf{Common Randomness:} $P_0,P_1$ hold random shares of 0 over $\mathbb{Z}_L$, denoted by $u_0$ and $u_1$ resp. \\
    \parbox{\textwidth}{\begin{enumerate}
        \item $P_0, P_1, P_2$ run $\prod _{\sf DReLU}(\{P_0, P_1\},P_2)$ with $P_j,j\in\{0,1\}$ having input $\langle a - 6 \rangle _j ^L$ and $P_0, P_1$ learn $\langle \alpha \rangle _0 ^L$ and $\langle \alpha \rangle _1 ^L$, resp.
        
        \item $P_0, P_1, P_2$ call $\prod _{\sf MatMul}(\{P_0, P_1\},P_2)$ with  $P_j,j\in\{0,1\}$ having input $(\langle \alpha \rangle _j ^L, \langle 6 - a \rangle _j ^L)$ and $P_0, P_1$ learn $\langle c \rangle _0 ^L$ and $\langle c \rangle _1 ^L$, resp.
        
        \item $P_0, P_1, P_2$ run $\prod _{\sf DReLU}(\{P_0, P_1\},P_2)$ with $P_j,j\in\{0,1\}$ having input $\langle a \rangle _j ^L$ and $P_0, P_1$ learn $\langle \beta \rangle _0 ^L$ and $\langle \beta \rangle _1 ^L$, resp.

        \item $P_0, P_1, P_2$ call $\prod _{\sf MatMul}(\{P_0, P_1\},P_2)$ with  $P_j,j\in\{0,1\}$ having input $(\langle \beta \rangle _j ^L, \langle a + c \rangle _j ^L)$ and $P_0, P_1$ learn $\langle d \rangle _0 ^L$ and $\langle d \rangle _1 ^L$, resp.
        
        \item For $j \in \{0,1\}$, $P_j$ outputs $\langle d \rangle _j ^L + u_j$.
    \end{enumerate}}% 
    \vspace{-0.5cm}
\label{alg:relu6_algo}
\end{mdframed}

\end{document}